\documentclass[pdflatex,sn-mathphys-num]{sn-jnl}% Math and Physical Sciences Numbered Reference Style 
\usepackage{graphicx}%
\usepackage{multirow}%
\usepackage{amsmath,amssymb,amsfonts}%
\usepackage{amsthm}%
\usepackage{mathrsfs}%
\usepackage[title]{appendix}%
\usepackage{xcolor}%
\usepackage{textcomp}%
\usepackage{manyfoot}%
\usepackage{booktabs}%
\usepackage{algorithm}%
\usepackage{algorithmicx}%
\usepackage{algpseudocode}%
\usepackage{listings}%

\theoremstyle{thmstyleone}%
\theoremstyle{thmstyletwo}%

\theoremstyle{thmstylethree}%

\begin{document}

%
%He had a weighing scale to compare two things together kind of like MMD which compares two things together
%
\title[Article Title]{HistoKernel: Whole Slide Image Level Maximum Mean Discrepancy Kernels for Pan-Cancer Predictive Modelling}


\author*[1]{\fnm{Piotr} \sur{Keller}}\email{Piotr.Keller@warwick.ac.uk}
\equalcont{First Author.}

\author[1]{\fnm{Muhammad} \sur{Dawood}}\email{Muhammad.Dawood@warwick.ac.uk}

\author[1,2]{\fnm{Brinder} \spfx{Singh} \sur{Chohan}}\email{b.chohan@nhs.net}

\author[1]{\fnm{Fayyaz} \spfx{ul Amir Afsar} \sur{Minhas}}\email{Fayyaz.Minhas@warwick.ac.uk}

\affil*[1]{\orgdiv{Tissue Image Analytics Centre}, \orgname{University of Warwick}, \orgaddress{\city{Coventry }, \postcode{CV4 7AL}, \country{United Kingdom}}}

\affil*[2]{\orgdiv{Department of Cellular Pathology}, \orgname{Royal Derby Hospital}, \orgaddress{\city{Derby}, \postcode{DE22 3NE}, \country{United Kingdom}}}

% commented
\abstract{Machine learning in computational pathology (CPath) often aggregates patch-level predictions from multi-gigapixel Whole Slide Images (WSIs) to generate WSI-level prediction scores for crucial tasks such as survival prediction and drug effect prediction. However, current methods do not explicitly characterize distributional differences between patch sets within WSIs. We introduce HistoKernel, a novel Maximum Mean Discrepancy (MMD) kernel that measures distributional similarity between WSIs for enhanced prediction performance on downstream prediction tasks.

Our comprehensive analysis demonstrates HistoKernel's effectiveness across various machine learning tasks, including retrieval (n = 9,362), drug sensitivity regression (n = 551), point mutation classification (n = 3,419), and survival analysis (n = 2,291), outperforming existing deep learning methods. Additionally, HistoKernel seamlessly integrates multi-modal data and offers a novel perturbation-based method for patch-level explainability. This work pioneers the use of kernel-based methods for WSI-level predictive modeling, opening new avenues for research. Code is available at 
 \href{https://github.com/pkeller00/HistoKernel}{https://github.com/pkeller00/HistoKernel}.}

\keywords{Computational Pathology, Mutation Prediction, Image Retrieval, Drug Sensitivity, Survival Analysis}



\maketitle

% \linenumbers
\section{Introduction}\label{sec1}
Advances in Computational Pathology (CPath) have demonstrated that machine learning can effectively utilize Whole Slide Images (WSIs) to address various clinically relevant problems~\cite{zhang2019pathologist,tolkach2020high}. Due to their large size and high memory demands, WSIs are divided into smaller patches for model training using weakly supervised approaches to perform patch-level predictions. These are then aggregated into a single slide-level prediction. The development of patch-level foundation models has further increased the need for efficient aggregation techniques~\cite{xu2024whole}.

Three main branches of aggregation techniques have emerged~\cite{bilal2023aggregation}. Heuristic methods generate predictions for each patch independently and aggregate them using fixed statistics, such as majority voting~\cite{bilal2021development,eastwood2023malignant}. These methods assume equal contribution from all patches, which fails to capture complex relationships and dependencies, and can lead to sensitivity to outliers and the overshadowing of significant patches if they are in the minority. Clinically driven aggregations use domain-specific rules, like calculating overall WSI tumor cell percentage based on patch-level counts~\cite{bilal2023aggregation}. While interpretable, they require established clinical rules, limiting their applicability. Data-driven methods, including attention mechanisms and graph neural networks (GNNs)~\cite{liang2023deep}, can weigh patches based on relevance and context, but their learned parameters are task-specific, making them less generalizable without extensive retraining, and are computationally intensive.

Crucially, existing aggregation approaches cannot explicitly quantify the differences or similarities in the multi-variable distributions of patches in two WSIs, limiting their effectiveness. This gap is addressed by our novel WSI-level Maximum Mean Discrepancy (MMD) kernels. These kernels act as statistical tests to determine if two WSIs, modeled as sets of patches, are drawn from different distributions by comparing infinite statistical moments~\cite{keller2023maximum}. This capability allows us to define pairwise similarity between WSIs, facilitating numerous downstream tasks such as visualization, clustering, regression, classification, and survival analysis~\cite{binder2021morphological,yoshida2018automated}.

% commentedz
\begin{figure}[!b]
\centerline{\includegraphics[width=\columnwidth]{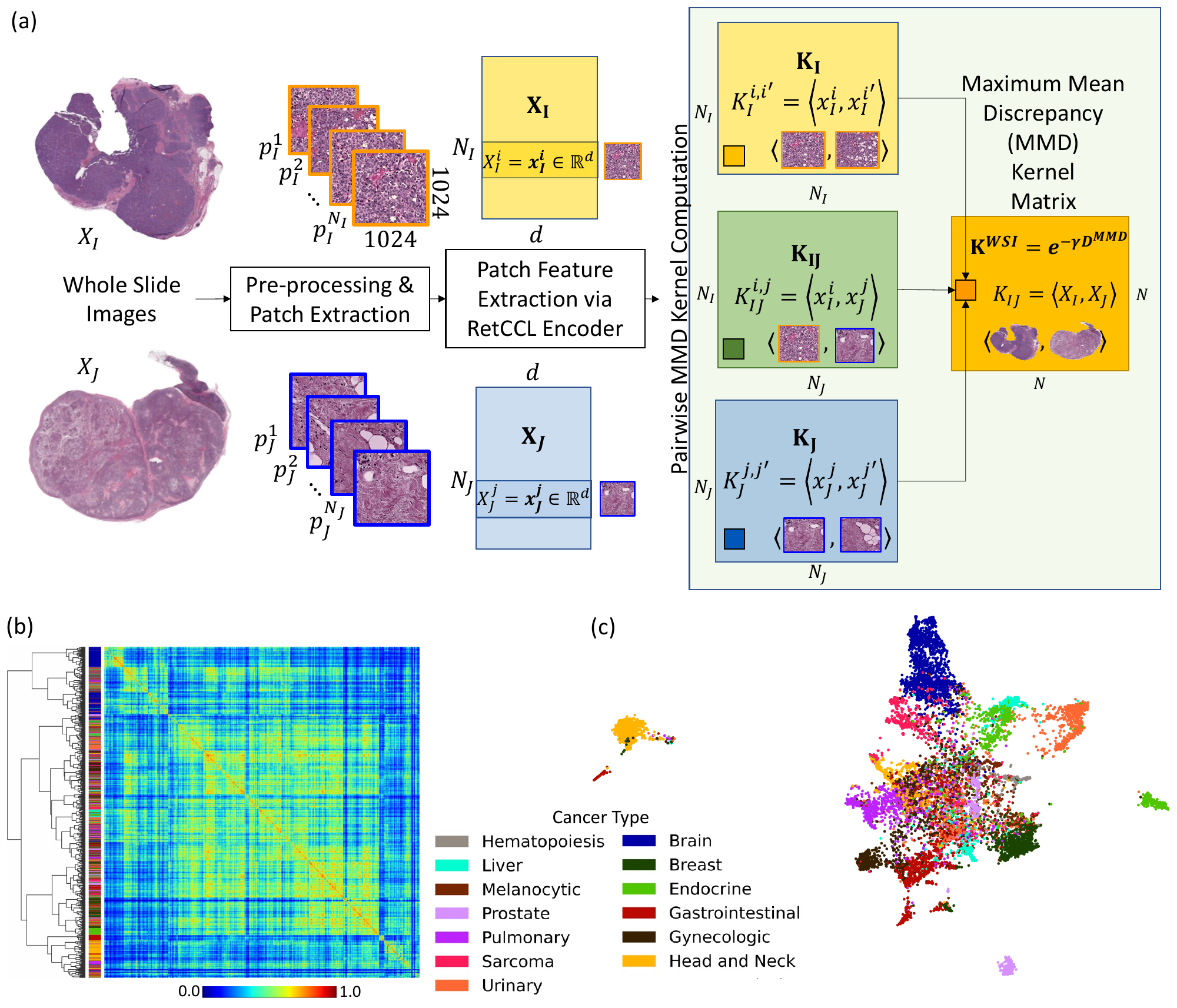}} 
\caption{(a) Workflow of Whole Slide Image (WSI) level Maximum Mean Discrepancy (MMD) kernel computation. Each WSI is processed by extracting $1024\times1024$ patches after applying a tissue mask~\cite{pocock2022tiatoolbox}. Patch features are then extracted using a feature encoder, and Equation~\ref{eq:MMDNew} computes the pairwise MMD kernel based on these features. This results in a pairwise MMD kernel matrix for the dataset with entries $K^{WSI}_{IJ}$  showing the similarity between any two WSIs. The pairwise MMD kernel matrix for all $n = 12,186$ WSIs in TCGA resulting in more than 74 million WSI pairs is shown in (b), with each entry representing WSI similarity (blue to red in intensity). Unsupervised hierarchical clustering of the kernel matrix reveals distinct clusters corresponding to cancer types (see legend), indicating the kernel captures expressive and meaningful information despite the method's unsupervised nature. Unsupervised UMAP visualization in (c) further corroborates this by showing meaningful clustering, with each dot representing a WSI. The WSI-level MMD kernel matrix is then used for downstream tasks such as classification, regression, retrieval and survival prediction with mutltimodal data integration.}
\label{fig:kernelVis}
\end{figure}

\section{Results}\label{sec2}
At its core, the proposed approach allows calculation of the degree of similarity between two WSIs in the form of a kernel function (henceforth called HistoKernel) based on patche feature embeddings. These patch-level feature embeddings can be obtained from any pre-trained embedding or foundation model (See Methods:~\ref{mmdExplanation} and Fig~\ref{fig:kernelVis} part (a)). Intuitively, if the kernel score between two WSIs is high then, with high probability, patches in these WSIs come from the same underlying distribution. Conversely, if the kernel score between two WSIs is small then that implies there is a substantial difference between the statistical distributions of patches comprising the two WSIs. This provides a principled way for identifying relationships between WSIs and allows the use of any kernel-based approaches for modelling downstream prediction tasks. Furthermore, for a given dataset, the kernel is computed only once and can be used for several tasks without re-computation leading to significant computational savings. We emphasise the kernel computation is purely unsupervised, no target labels are required for computing HistoKernel. 

Below, we demonstrate the versatility and effectiveness of HistoKernel across multiple tasks in CPath such as visualization and clustering, WSI retrieval, regressing cancer drug sensitivity, classification of point mutation classifications and multi-modal data integration for survival analysis.

\subsection{Visualization and Clustering}
Fig~\ref{fig:kernelVis} (b) shows HistoKernel for the entire TCGA dataset comprising more than 74 million pairs of WSIs. The degree of similarity between any two WSIs is expressed as a single number leading to the interpretation of the kernel matrix as an affinity matrix that can be used for hierarchical clustering or two-dimensional visualization through UMAP (see Fig~\ref{fig:kernelVis} (b) and (c)). These visualizations clearly show WSIs of the same cancer type are clustered closer to each other in comparison to those from other types. Thus, HistoKernel is able to capture meaningful similarities between WSIs based on cancer type despite the fact that no cancer type labels or any other type of target labels are used in kernel computation. The kernel matrix visualization provides useful insights into similarities and differences in WSIs of different cancer types which typical patch-based approaches cannot do. For example, the UMAP plot shows some cancer types are very distinguishable such as brain and prostate whilst others are more closely clustered such as liver and endocrine tumours. It also allows for identification of cancer sub-populations such as the two predominant sub-clusters in endocrine tumours. Investigating the reasons for these patterns can help broaden our understanding of cancer. Dataset visualization can also be used for identifying potential noise or outliers, allowing interactive quality control in a CPath pipeline. 

\subsection{Whole Slide Image Retrieval}
We utalise HistoKernel for WSI retrieval. Here, we are interested in retrieving the top $k$ most similar WSIs for a given query image, $X_q$. Effective retrieval can be used to aid in clinical diagnosis for new patients based on diagnosis of similar patients. It may also help train new histopathologists by showing them similar images with the same diagnosis to help identify common histology patterns such as gland shape often used for prostate cancer grading~\cite{bostwick1994grading}.

To perform WSI retrieval we directly query HistoKernel matrix to find the most similar images (see Methods:~\ref{methodWSIRet}). We compared performance of the proposed approach with current State of the art (SOTA), RetCCL, under the same validation protocol~\cite{retccl}.

Table~\ref{RetrivalResults} shows the majority vote at the top five search results ($mMV@5$) achieved by HistoKernel and SOTA~\cite{retccl} for each cancer sub-type. The $mMV@5$ measures the percentage of samples in test set for which the majority cancer sub-type in the $top-k$ retrieved samples is the same as the query image. We see HistoKernel consistently outperformed RetCCL, 12\% higher $mMV@5$ macro-average across all sub-types. Further, HistoKernel's results are statistically significantly higher than comparative method (Wilcoxon paired signed rank test found $p \ll 0.01$). Thus, overall we can be confident HistoKernel has significantly outperformed the SOTA for WSI retrieval. 

% commented
\begin{table}[!b]
\caption{The results of cancer sub-type retrieval. The table shows the majority vote at the top $5$ search results ($mMV@5$) achieved by HistoKernel and comparative method~\cite{retccl} for each cancer sub-type considered in the TCGA. The $mMV@5$ metric measures the percentage of samples in test set for which the majority cancer sub-type in the $top-k$ retrieved samples is the same as the query image. From this table it is clear that HistoKernel significantly outperformed the comparative method for majority of cancer types suggesting HistoKernel has learned a more meaningful representation than the comparative method.}
\label{RetrivalResults}
\setlength{\tabcolsep}{3pt}
\begin{tabular*}{\textwidth}{@{\extracolsep\fill}lcccc}
\toprule%
WSI Type & \#WSIs & \#Patients & mMV@5 &  \\ 
\cmidrule{4-5}
& & & RetCCL & HistoKernel\\
\hline
\textbf{Pulmonary} & & & & \\
LUAD & 531 & 468 & 53.86& \textbf{73.44} \\
LUSC &512 &478 & 75.20& \textbf{88.28} \\
LUSC &512 &478 & 75.20& \textbf{88.28} \\
MESO & 73& 67& 41.10& \textbf{46.58} \\
\hline
\textbf{Urinary} & & & & \\
BLCA & 457 & 386&91.68& \textbf{98.47} \\
KIRC & 519 & 513& 86.32& \textbf{92.68} \\
KICH &121& 109&49.59& \textbf{74.38} \\
KIRP & 297& 273& 50.51& \textbf{81.48} \\
\hline
\textbf{Gastrointestinal} & & & & \\
COAD & 551 & 452 & \textbf{90.20} & 85.48\\
ESCA & 157 & 155 & 50.96 & \textbf{74.52} \\
READ & 183 & 161 & 5.46 & \textbf{12.02} \\
STAD & 400 & 375 & 69.00 & \textbf{73.75} \\
\hline
\textbf{Melanocytic} & & & & \\
UVM & 68 & 68 & 92.65 & \textbf{100.0}  \\
SKCM & 473 & 431 & 98.73 & 98.73 \\
\hline
\textbf{Brain} & & & & \\
GBM & 842 & 378 & 68.76 & \textbf{90.50} \\
LGG & 843 & 491 & \textbf{92.41}  & 80.66\\
\hline
\textbf{Liver} & & & & \\
CHOL & 39 & 39 & 5.13 & \textbf{35.90} \\
LIHC & 372 & 364 & \textbf{97.58} & 96.51 \\
PAAD & 209 & 183 & 96.17 & \textbf{99.04} \\
\hline
\textbf{Gynecologic} & & & & \\
UCEC & 566 &505 & \textbf{93.64}& 89.22 \\
CESC & 279 & 269 & 62.37& \textbf{84.23}\\
UCS & 72& 45&29.17 & \textbf{59.72}\\
OV & 107& 106&41.12&\textbf{83.18} \\
\hline
\textbf{Endocrine} & & & & \\
ACC  & 227 & 56 & 85.02 & \textbf{92.95}\\
PCPG & 192 & 173 & 76.56 & \textbf{91.15}\\
THCA & 447 & 437 &96.42& \textbf{97.09}\\
\hline
\textbf{Prostate/Testis} &  &  & & \\
TGCT & 215 & 138& 93.49 & \textbf{97.21} \\
PRAD & 379 & 343 & 97.89 & \textbf{98.68} \\
\hline
\textbf{Hematopoiesis} & & & & \\
DLBC & 36 & 36 & 27.78 & \textbf{63.89} \\
THYM & 157 & 107 & 98.09 & \textbf{99.36} \\
\end{tabular*}
\footnotetext{The macro-average mMV@5 of RetCCL and HistoKernel are 69.55 and 81.35, respectively.}
\end{table}

\subsection{Regressing Cancer Drug Sensitivities}
Next, we demonstrate HistoKernel's capability for regression tasks by predicting cancer drug sensitivity from WSIs. We want to predict if a patient will respond to a drug using only a WSI. The target drug sensitivity values are inferred as real numbers by aligning patients gene expression profile with cell line expression profile for which drug sensitivity values of drugs exist~\cite{gruener2021facilitating}. This is an important task as drug sensitivities can be used to optimise a patient’s treatment, many studies supporting this approach~\cite{kidwai2022pharmacogenomics}. Predicting drug sensitivities just from images is particularly interesting as histopathological assessment is considered routine and may be more readily available than genetic molecular tests which have high costs.

We can utilise the pre-computed HistoKernel to train a Support Vector Regressor (SVR) to perform predictions. We compare performance of HistoKernel with the existing SOTA approach in this domain by Dawood et al.'s SlideGraph$^\infty$ under the same evaluation protocol~\cite{dawood2024cancer} with spearman rank correlation between true and predicted sensitivities for each drug as a performance metric.

\begin{figure}[!t]
\centerline{\includegraphics[width=\columnwidth]{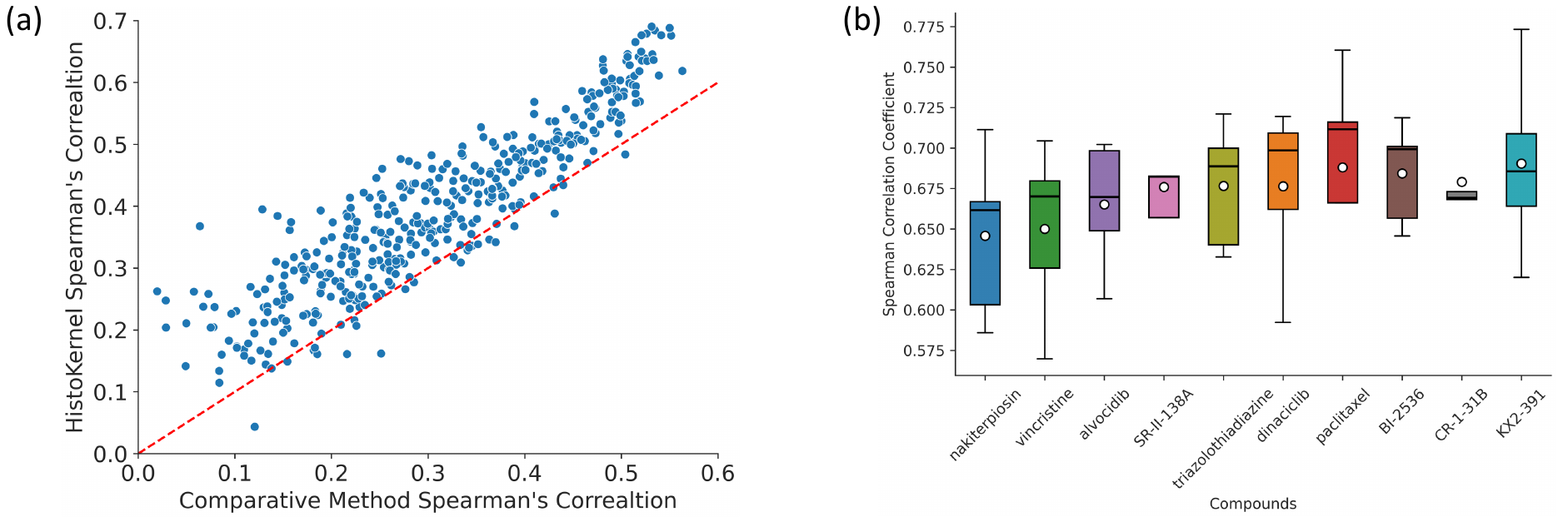}}
\caption{Results of regressing drug sensitivities in breast cancer for Histokernel and the SlideGraph GNN comparative method~\cite{dawood2024cancer}. In (a) each blue dot in the plot represents the mean Spearman correlation coefficient (SCC) achieved for a given compound by our Histokernel (y-axis) and the comparative method (x-axis). This plot clearly shows that HistoKernel has outperformed comparative method as it produced better drug sensitivity predictors for 407 out of 427 compounds. As such HistoKernel was better able to identify histopathology based patters associated with drug sensitivity compared to comparative method. In (b) we see the distribution of SCC across the 5 folds for the top 10 drugs predicted by HistoKernel. Boxes depict quartile values and the whiskers extend to 1.5 times the interquartile range. The black line and white dot represent the median and mean of the distribution respectively. Being able to identify drug sensitivity only from WSI with high SCC for drugs such as Paclitaxel means we can identify which patients would most likely benefit from Paclitaxel thus helping improve patient quality of care. Further we can see thse drugs have narrow distributions across the 5-folds strengthening the reliability of these models.}
\label{SCCRegressor}
\end{figure}

Fig~\ref{SCCRegressor} part (a) shows the comparison between Spearman correlation coefficient (SCC) of predicted sensitivity for HistoKernel (y-axis) and the comparative method (x-axis)~\cite{dawood2024cancer} for each compound. HistoKernel outperformed comparative method in terms of SCC for 94\% of compounds clearly showing its superiority. HistoKernel's results are statistically significantly higher than comparative method (Wilcoxon paired signed rank test found $p \ll 0.01$) indicating a genuine increase in performance that was not a result of chance. Finally, in Supplementary Figure 1 we see HistoKernel produced $1.5$ times as many statistically significant predictors (p-value $\leq1e^{-3}$) compared to comparative method indicating greater reliability of HistoKernel. Fig~\ref{SCCRegressor} part (b) shows the distribution of HistoKernel's top 10 models which can have numerous advantages for patients treatment success. For example, Vincristine (SCC=$0.65 \pm 0.05$) and Paclitaxel (SCC=$0.69 \pm 0.06$) are not effective for all patients~\cite{ponde2019progress}. Consequently, HistoKernel provides a potential mechanism for early identification of patients for alternative therapies leading to improved treatment efficacy and minimising side effects~\cite{wu2023proactive}. 

% commented

\subsection{Classification of Point Mutations}
\label{ResultsPointMut}

To illustrate the efficacy of HistoKernel for slide-level classification problems, we perform point mutation classification. Point mutations are genetic mutations that involve a change in a single nucleotide base within the DNA or RNA sequence~\cite{pointMutationDEF}. Numerous studies reported associations between various genetic mutations and tumour progression and therapeutic response, highlighting the tasks importance~\cite{fontanini1998evaluation}. WSI-based point mutation predictors can act as alternatives to genetic test leading to reduced turnaround times and costs~\cite{rusch2018clinical}.

HistoKernel was used as a precomputed kernel to train a Support Vector Machine (SVM) to perform gene point mutation classification. We compared performance of the proposed approach with the current SOTA under the same validation protocol~\cite{saldanha2023self}.

\begin{figure}[!t]
\centerline{\includegraphics[width=\columnwidth]{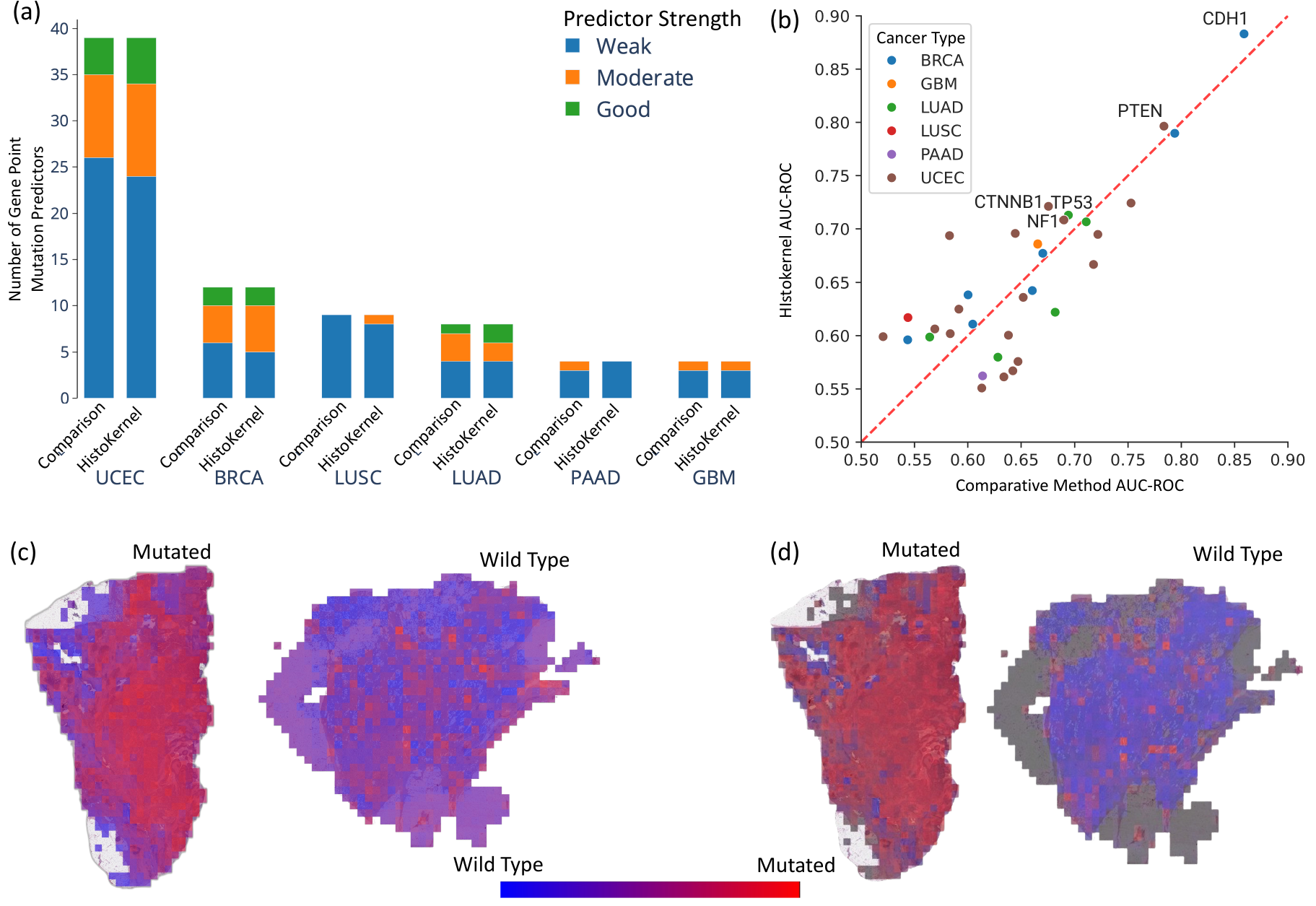}}
\caption{The results of gene point mutation prediction in six cancer types for HistoKernel and an SOTA attention-based neural network~\cite{saldanha2023self}. Part (a) shows the results of gene point mutation in terms of of AUC-ROC by HistoKernel and the SOTA~\cite{saldanha2023self}. Individual gene classifiers for each method are binned into 3 groups based on their strength where blue represents weak classifiers, orange are moderate classifiers and green are good classifiers. Here HistoKernel produces 9 good and 19 moderate predictors in contrast to SOTA which produced 7 good and 18 moderate. This suggests HistoKernel slightly outperforms the SOTA suggesting the two methods are comparable in capturing morphological patters associated with point mutations. This idea is strengthened by part (b) which shows the pairwise comparison of comparative method and HistoKernel for each gene where either approach achieved an AUC-ROC $\geq 0.6$ (34 genes). We see HistoKernel outperformed comparison for 19 out of 34 filtered genes suggesting it is a marginally better method for point mutation prediction. Part (c) the heatmap of HistoKernel for a TP53 mutated WSI and a wild type image is shown. Part (d) we show a comparison of comparative method's heatmap for the same slides as in part (c). These heatmaps show that our approach is identifying similar patterns as a well established method thus giving HistoKernel credibility in its predictions.}
\label{pointPredictionBar}
\end{figure}

Comparison results in Fig~\ref{pointPredictionBar} part (a) show the number of genetic point mutations for a given cancer type that were predicted with high, moderate or weak confidence by HistoKernel and current SOTA~\cite{saldanha2023self}. We see HistoKernel produced 48 weak classifiers whilst SOTA produced 51 thus suggesting HistoKernel performed slightly better even without any training of the underlying patch feature extractor. Further, in Fig~\ref{pointPredictionBar} part (b) we see direct pairwise comparison of each gene where again HistoKernel performed marginally better (outperforming SOTA for 56\% of genes). Overall, this suggests HistoKernel is comparable or even slightly better than current SOTA which shows its ability in a clinically relevant classification task. Two genes to note are CDH1 in breast adinocarcinoma (AUC-ROC $0.88 \pm 0.02$) and PTEN (AUC-ROC $0.80 \pm 0.04$) in endometrial cancer for which HistoKernel outperformed SOTA. Strong predictors for these genes can help identify patients who may benefit from targeted therapy~\cite{m2014therapeutic,citrin2018treatment,m2014therapeutic}. Finally, Fig~\ref{pointPredictionBar} part (c) and (d) show heatmaps of HistoKernel's and SOTA's~\cite{saldanha2023self} patch-level predictions respectively for TP53 point mutation prediction in breast cancer. To generate HistoKernel's patch-level predictions we utilize our novel patch sensitivity method (See Methods:~\ref{Explainability}). These visualisations are promising as HistoKernel is identifying similar regions to current SOTA method~\cite{saldanha2023self} thus giving HistoKernel predictions credibility.

% commented

\subsection{Survival Analysis}
\label{sec:survAnalysisRes}
Thirdly, we use HistoKernel for survival analysis. Here we use WSIs to predict the expected time until a clinically important event occurs, for example, disease progression or death, given other patients survival data. Accurate survival predictions can be used by pathologists to make data-driven decisions about best treatment course, improving overall patient care. Using WSI-based deep learning we can reduce human subjectivity associated with analysing tissue visually~\cite{bo2017intra} as well as the identification of significant prognostic patterns.

HistoKernel can be utilised as a precomputed kernel for a kernelized survival SVM (KSSVM)~\cite{fastTrainingSurvivalAnalysis} allowing us to perform survival analysis. We compare the performance of HistoKernel with a comparative model by Mackenzie et al.'s (a survival GNN) under the same evaluation protocol~\cite{mackenzie2022neural}.

\begin{figure}[!t]
\centerline{\includegraphics[width=\columnwidth]{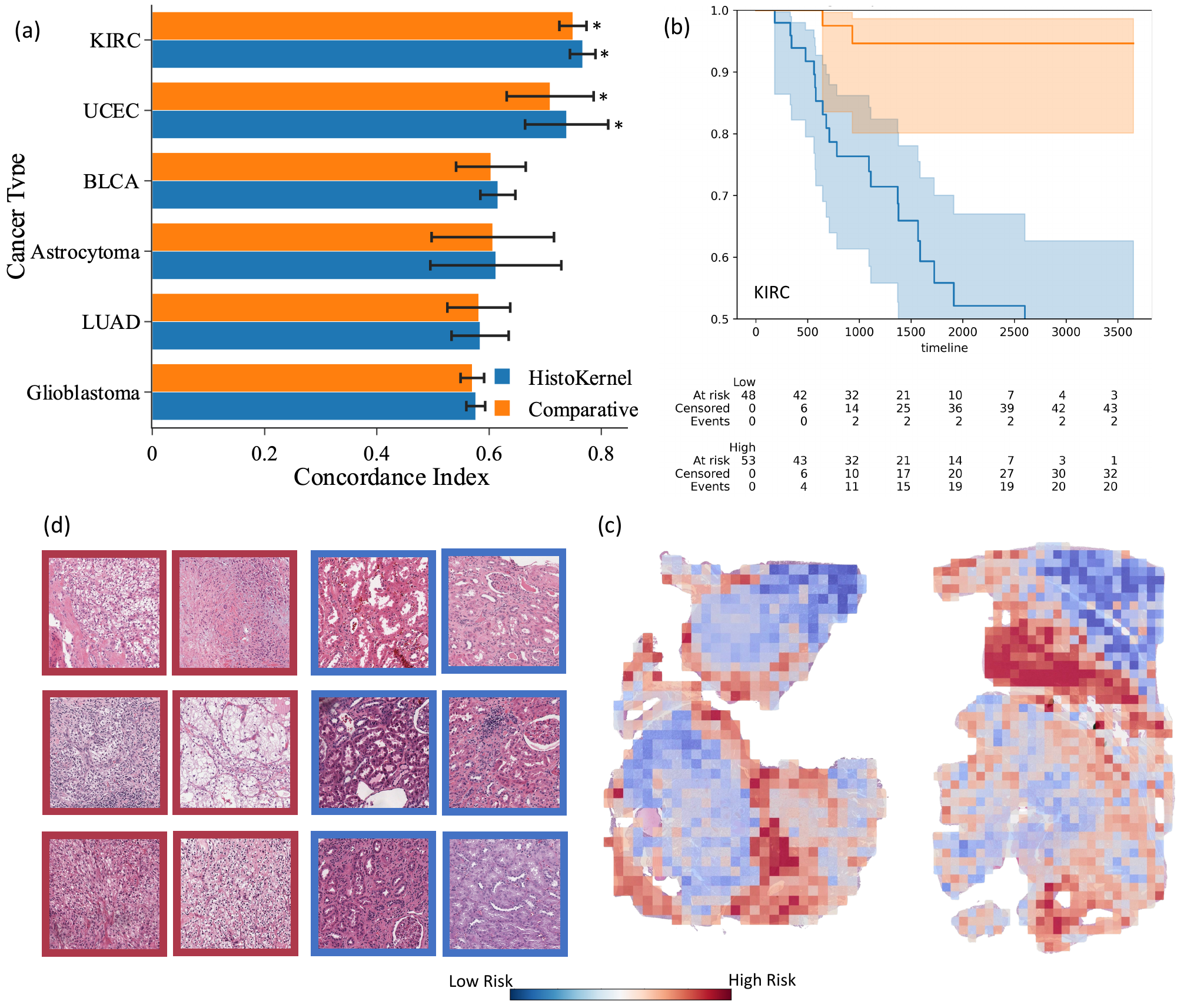}}
\caption{(a) Results of survival modelling achieved by Histokernel and a GNN based comparative method~\cite{mackenzie2022neural} for six different cancer types. Kidney Renal Clear Cell Carcinoma (KIRC), Uterine Corpus Endometrial Carcinoma (UCEC), Astrocytoma, Glioblastoma, Bladder Urothelial Carcinoma (BLCA) and Lung Adenocarcinoma (LUAD) are considered in this study. For each cancer type, we report both model's C-index along with the respective standard deviation (the black line) across the five folds. The asterisk indicates that the results are statistically significant, p-value $<0.05$. This bar chart clearly shows that HistoKernel outperformed comparative method as it achieved a higher C-index for all cancers considered. This suggests HistoKernel predictors are better for these cancer types and have greater potential for clinical applications than comparative method. Part (b) shows the Kaplan-Meier survival curve  of HistoKernel for Kidney Renal Clear Cell Carcinoma ($p=0.000$). High-risk and low-risk patients are represented by blue and orange respectively. From this figure it is clear that samples have been successfully stratified into high and low risk patients. Such a stratification may be useful to doctors in deciding the best course of treatment for a given patient indicating potential for clinical applications. In part (c) you can see a few representative patches from high and low risk regions. Such patches provide insight into HistoKernels patterns in the entire dataset without the need for a pathologist to evaluate a large number of WSIs. For a deeper analysis in (c) slide-level heatmaps of WSIs associated with the representative patches. Blue regions indicate low risk areas whereas red regions indicate high risk areas. Such patch-level predictions provide explainability to the model and help doctors trust the predictions more in a clinical setting.}
\label{survivalKMPlots}
\end{figure}

The performance over six different survival prediction tasks is compared using Concordance index (C-index) which measures the extent to which the rankings of patients generated by the model align their actual survival times. These scores are reported for both methods along with the standard deviation across the five folds with the associated p-value in Figure~\ref{survivalKMPlots} (a). This shows that HistoKernel outperforms the comparative method for all considered cancer types. The Kaplan-Meier plots for Kidney Renal Clear Cell Carcinoma (KIRC) produced by HistoKernel is shown in Fig~\ref{survivalKMPlots} (b). A clear separation between high and low risk patients is shown which medical practitioners can use to select a patients' appropriate treatment. To provide explainability to HistoKernel we utilised our perturbation-based approach. To do this we assign all patches a patch-level score and then perform clustering to identity representative poor and good prognosis patches as shown in Fig~\ref{survivalKMPlots} (c). A trained pathologist identified clear differences between morphology of the two patch sets. Specifically, patches predicted to have good prognosis all show benign tissue (mainly renal parenchyma). Conversely, those predicted to have poor prognosis mostly contain clear cell renal cell carcinoma showing chronic inflammatory cells, mainly lymphocytes. To provide a slide-level perspective a pathologists took a deeper look at 5 WSIs that were associated with 5 of these representative patches. These slide-level visualisations in Fig~\ref{survivalKMPlots} (d) supported the idea that low risk regions mostly showed benign tissue. High risk regions contained various patterns such as the pressure effect of tumours, tumour interface to stroma and tumour in the vicinity of large caliber vessels. This improves trustworthiness of Histokernel as it has picked up clinically relevant information that has been associated with prognosis in current literature~\cite{nilsson2020features,mattila2022prognostic}.

\subsection{Multi-Modal Integration with Kernels}
Finally, we show the ease of integrating multiple data modalities with HistoKernel. As a proof of concept, we demonstrate results for multi-modal survival analysis in BRCA. Here we want to combine information from different data sources such as histological image data and transcriptomics in order to capture complementary information about patient survival. 

We build on Multiple Kernel learning to combine the pre-computed HistoKernel with a kernel based on patient transcriptomic profiles or topics (see methods:~\ref{sec:methodMultiLearning})~\cite{gonen2011multiple}. We can then utilise this combined kernel as a precomputed kernel for a KSSVM (as explained in Results:~\ref{sec:survAnalysisRes}). Specifically, we use additive and multiplicative combinations of kernels from the two modalities. 

\begin{figure}[!t]
\centerline{\includegraphics[width=\columnwidth]{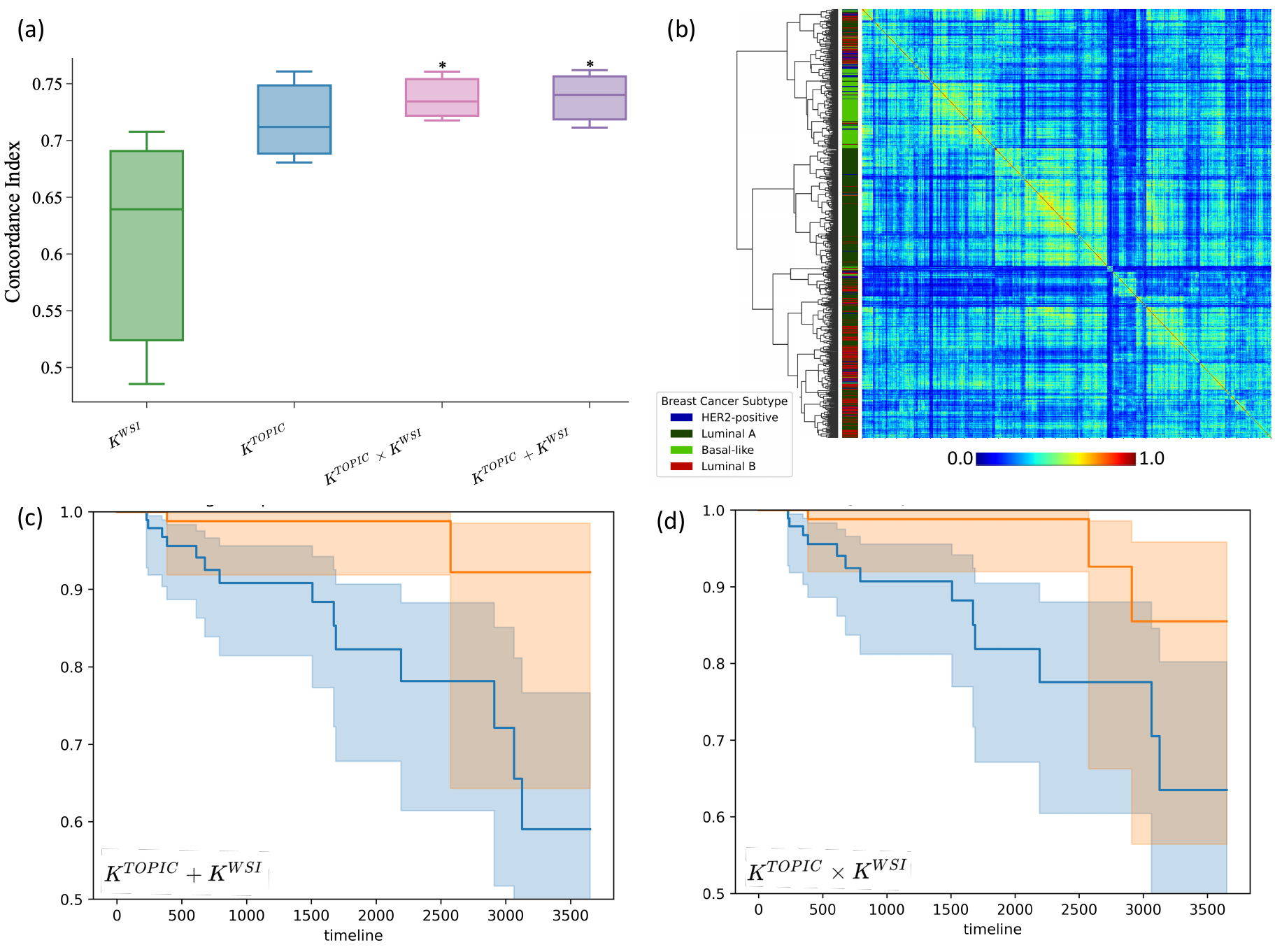}} 
\caption{(a) Results of multi-modal survival analysis for breast cancer with two multi-modal kernels being utilised as pre-computed kernels for KSSVM. The base single-modal kernels are a  morphological Histokernel ($K^{WSI}$) as well as a transcriptomic topic-based kernel ($K^{TOPIC}$). These base kernels are combined via addition and multiplication in order to generate two new kernels, $K^{TOPIC}+K^{WSI}$ and $K^{TOPIC}\times K^{WSI}$ respectively. From this plot it is clear that multi-modal kernels achieve higher concordance index indicating that they are able to better capture patters associated with patient survival. Further, these combined kernels are more stable as they have a smaller standard deviation across the three folds, indicating better consistency. Finally, only the combined kernels are statistically significant (shown by an asterix) further showing their superiority. Focusing on the $K^{TOPIC}+K^{WSI}$  which achieved the highest C-index we can visualise it to get a deeper understanding. In (b) we see a heatmap of this kernel matrix such that each entry represents patient similarity (blue to red in intensity). Unsupervised hierarchical clustering of this kernel matrix reveals clear clusters corresponding to breast cancer subtypes (see legend). This suggests the kernel is able to capture expressive information about breast cancer, despite its generation being completely unsupervised. Finally, part (c) and (d) further corroborates the benefit of a multi-modal approach as the Kaplan–Meier curves of both multi-modal kernels show a good separation between high and low risk patients, further indicating that these kernels were able to learn meaningful patterns associated with survival.}
\label{fig:MMLFigure}
\end{figure}

Figure~\ref{fig:MMLFigure} (a) shows the results of survival prediction in terms of C-index for BRCA for two multi-modal kernels. We see that both multi-modal kernels perform better than their single-modal counterparts. A clear increase in C-index as well as a smaller standard deviation clearly shows a multi-modal approach better captures patters associated with survival whilst being more robust. Part (c) and (d) support this idea as Kaplan–Meier curves produced for both multi-modal kernels show a clear separation between high and low risk patients. Further analysis of the best performing $K^{TOPIC} + K^{WSI}$ kernel shows that unsupervised hierarchical clustering of the kernel (b) finds clear clusters forming with respect to BRCA subtypes. For example, basal-like breast cancer forms a clear cluster and is known to be associated with worse prognosis~\cite{lachapelle2011triple}. These results show the ease with which imaging and transcriptomic data can be integrated. 

% commented

\subsection{Time Complexity}
For kernel computation we computed similarity between 12,186 WSIs, resulting in 74,243,205 WSI pairs. Calculating a single pairwise computation varies depending on number of patches but on average took 4 miliseconds. Thus, for all pairs the distance matrix computation requires 3.4 days on single GPU. This can be accelerated with parallelisation and by preloading WSI feature vectors into the GPUs.

After kernel computation the same precomputed kernel can be utilised in numerous kernel-based algorithms for downstream tasks. By utilising a precomputed kernel training is extremely fast in contrast to traditional deep learning approaches. For drug sensitivity prediction training all 2135 models (427 drugs with 5-fold cross validation) including hyper-parameter optimisation took 15 minutes on a machine with 32GB of RAM and an Intel Core i5-13500 processor. On the same machine, point mutation prediction computation took 6 minutes for all 304 models (76 genes with 4-fold cross validation) including hyper-parameter optimisation. WSI retrieval took 10 minutes for retrieving top 5 images for all 9,362 WSIs considered. Finally, survival analysis for the 30 models (6 sites with 5-fold cross validation) took 10 minutes. These times are a significant improvement over existing approaches. 

\section{Discussion}
We presented a novel MMD-based kernel approach that quantifies similarity between WSIs. By computing MMD between WSIs represented as sets of patch-level features we generate a single kernel for entire dataset where the similarity between any two given WSIs is expressed by a single number. This kernel is only computed once for the entire dataset, after which it can be used for a variety of downstream tasks. Avoiding long retraining times per task makes HistoKernel very fast and flexible. Nevertheless, HistoKernel captures meaningful information as it is comparable to SOTA approaches for numerous fundamental tasks. To strengthen this claim, our perturbation-based patch-level explainer found logical patterns in HistoKernel’s predictions, highlighting the potential for clinical applications.

Firstly, we visualised HistoKernel. Fig~\ref{fig:kernelVis} part (a) shows that numerous cancers like brain form clear clusters, supporting the idea that HistoKernel is finding patterns in data. This clustering is not perfect potentially due to tumour heterogeneity. Fig~\ref{fig:kernelVis} part (b) further support idea of cancers clustering. Interestingly, both clustering methods coincide with the cancer types that are most distinguishable. It may be beneficial to study underlying reasons for this. It seems HistoKernel is capturing meaningful patterns in the data. It is unlikely HistoKernel is picking up stain variations or possible artefacts as these have been accounted for during feature extraction and tissue segmentation. However, other confounding variables may exist.

We investigated WSI retrieval where HistoKernel significantly outperformed SOTA for majority of sites\footnote{The RetCCL results in this paper are different from those originally reported due to a different subset of TCGA and tissue mask algorithm being used~\cite{pocock2022tiatoolbox}}, 12\% higher macro-average. A strong retrieval method is useful for diagnosis of new patients, based on similarity with patients with existing diagnosis. Further, retrieval can be used to discover new patterns, for example identifying genetic patterns in patients who have high image-based similarity. Overall, HistoKernel has outperformed SOTA for retrieval without any processing of the kernel. This strengthens the argument that HistoKernel has found meaningful patterns.

Next, we looked at the regression task of predicting drug sensitivities. HistoKernel significantly outperformed SOTA, achieving higher SCC for 94\% of compounds. For Vincristine and Paclitaxel HistoKernel achieved highest SCC. Knowledge of patient sensitivity to these compounds can reduce serious side effects such as Vincristine neurotoxicity, improving treatment efficacy and patient prognosis~\cite{ponde2019progress,wu2023proactive}. Overall, HistoKernel has significantly outperformed comparative method for regression task without recomputing the kernel.

% For PTEN HistoKernel may be a strong predictor as this mutation has been associated with various morphological patterns which can be picked up by image-based classifiers~\cite{gbelcova2022pten}.
Using the same kernel we explore classification (point mutation prediction). HistoKernel was comparable to SOTA, outperforming it for 56\% of genes. Particularly, HistoKernel achieved higher AUC-ROC than comparative method for CDH1 and PTEN in BRCA and UCEC respectively, for which targeted therapies exist~\cite{m2014therapeutic,citrin2018treatment}. This may improve patient care by identifying effective therapies for patients quickly and cheaply (in contrast to genetic tests). We also compare patch-level heatmaps of both HistoKernel and SOTA for TP53 prediction. These highlight similar areas suggesting HistoKernel learned similar patterns to a previously well-established method, giving it credibility. However, 42 genes were still unpredictable for both methods. This may be because these genes may not be associated with significant morphological patterns. Thus, HistoKernel is comparable to SOTA and has managed to identify meaningful information for a clinically relevant classification task.

% HistoKernel is better for BLCA and LUAD than comparative method however C-index is still low so it is unlikely that HistoKernel can be used without expert monitoring in a clinical setting for these cancers.
Next, we consider a ranking problem, survival analysis. HistoKernel has outperformed comparative method for all cancer types considered. However, results are statistically significant only for KIRC and UCEC. This may be due to limited number of samples per cancer type. Kaplan-Meier plots curves of highest performing model for KIRC show that HistoKernel could separate patients into two statistically significantly different groups. Low-risk patients have significantly better survival probability over time compared to high-risk group. A good stratification is useful for doctors to decide the best treatment course, helping improve quality of care. Pathologist analysis of HistoKernel's representative patches showed it successfully identified low risk regions as benign tissue, mainly renal parenchyma (some showing fat, likely perinephric). For high risk patches most contained clear cell renal cell carcinoma, containing chronic inflammatory cells mainly lymphocytes. Further slide-level analysis of 5 WSIs associated with 5 representative patches supported the idea that low-risk regions were mostly benign tissue. The pathologist identified various patterns in HistoKernel high-risk regions of these WSIs. This included: tumour interface to stroma, tumour in the vicinity of large caliber vessels, pressure effect off tumours (pushed aside parenchyma and significant fibrosis and chronic inflammation) and areas of higher grade tumour (more eosinophilic and more prominent nuclei). This shows HistoKernel has successfully identified clinical information that current literature has shown to be associated with prognosis without direct annotations or specific labels (other than basic patient's survival information)~\cite{nilsson2020features,mattila2022prognostic}. This provides credibility to HistoKernel thus opening avenues for clinical applications. However, certain patterns need further investigation. For example, HistoKernel has identified Granulomas outside of the tumour itself (in the interring tissue, possible in the vicinity of sinus fat) as an important factor for worse prognosis. However, current literature is inconclusive of Granulomas importance due to its rarity in KIRC. Overall, HistoKernel outperformed comparative method for numerous cancers in a fundamentally different problem and showed clinically relevant patterns whilst utilising the same kernel.

Finally, we demonstrate potential for multi-modality integration with multi-kernel learning (MKL) via multi-modal survival analysis results. These show a multi-modal approach significantly outperformed single-modal variants. This substantial increase in performance and reduced standard deviation showed clear advantage of MKL. Furthermore, integration of two or more kernels is very easy, utilising basic arithmetic to combine kernels. This is supported by Kaplan-Meier plots of the multi-modal kernels showing a good stratification between high and low risk patients. Finally, unsupervised clustering of the strongest performing kernel showed clear clusters forming with respect to breast cancer subtypes suggesting that multi-modal kernel contains meaningful information~\cite{lachapelle2011triple}. It is possible the multi-modal kernel is utilising this information as current literature shows breast cancer subtype is associated with survival. Overall, a multi-modal approach is very easy to integrate into existing pipelines.

% % commented
\section{Methods}\label{sec4}
\subsection{Dataset}
For all the experiments we used WSIs of Formalin-Fixed paraffin-Embedded (FFPE) Hematoxylin and Eosin (H\&E) stained tissue section from The Cancer Genome Atlas (TCGA)~\cite{hoadley2018cell}. After excluding WSIs having missing baseline resolution information in total we end up  with 12,186 WSIs belonging to 9,374 patients. We used data of 32 caner sub-types spanning across 25 anatomic sites.

To obtain the target labels of patients point mutation status we downloaded mutation data of patients in 6-cancer types (endometrial, breast, lung adenocarcinoma, lung squamous cell carcinoma and glioblastoma) from cBioportal~\cite{cerami2012cbio}. Inline with previous work we curated cancer-related genes with at least 25 mutations of known significance with a matching WSI~\cite{saldanha2023self}. However, unlike previous work we relabelled TCGA brain samples to obtain a new glioblastoma sub-population based on modern classification standards~\cite{louis20212021} (2021 fifth edition WHO). This edition incorporated molecular features into the diagnostic decision tree for brain tumours thus causing a major change in the classification of samples. These reclassified labels for the TCGA were obtained from Zakharova et al.~\cite{zakharova2022reclassification}. Due to this relabelling and after excluding WSIs with missing baseline resolution from analysis we obtained 76 analysable genes in six cancer types.

For cancer drug sensitivity prediction to be consistent with comparative method, we used data of 1,098 patients with breast invasive carcinoma from TCGA (TCGA-BRCA). For these patients we acquired their estimated sensitivity to 427 experimental and FDA-approved compounds from a previously published work that has used genomic and drug response data of Cancer Cell Lines (CCL) data~\cite{drugComputation} to infer these numbers. Based on these labels the comparative method restricted the analysis to patients with gene expression-based imputed sensitivity scores for all compounds to get a total of 551 patients for analysis. The labels were also normalised using z-score normalisation to account for the different ranges of the drug scores.

For survival analysis, we restricted our analysis to kidney renal clear cell carcinoma (n=504), uterine corpus endometrial carcinoma (n=504), astrocytoma (n=219), glioblastoma (n=266), bladder urothelial carcinoma (n=372) and lung adenocarcinoma (n=426). To obtain labels for astrocytoma and glioblastoma we relabelled the brain Lower Grade Glioma and Glioblastoma Multiforme cohorts from the TCGA~\cite{louis20212021}. The Disease-Specific Survival (DSS) data for each patient was collected from the TCGA Pan-Cancer Clinical Data Resource (TCGA-CDR)~\cite{liu2018integrated}. Patients with missing DSS data or those with an event time less than or equal to zero were not used in the study.

For WSI retrieval, in line with comparative method, we looked at the 10 cancer source sites which have more than one sub-type associated with them (29 sub-types in total). This resulted in 9,324 WSIs belonging to 7,606 patients.

For Multi-Kernel Learning we obtained gene expression states (topics) from a previous method that utalised CoReX on normalised gene expression data~\cite{dawood2023data}. We restricted our analysis to breast adenocarcinoma as topics have only been computed for this cohort, resulting in 956 patients. Corresponding survival data was obtained from TCGA-CDR~\cite{liu2018integrated}.

\subsection{Pre-processing of WSIs}
For each WSI we first identify its viable tissue regions by using a U-Net-based tissue segmentation model from the TIAToolbox~\cite{pocock2022tiatoolbox}. This ensures any artefacts such as pen markings or tissue folds as well as background regions are removed. This generates a mask for each image with a score of one for tissue area and zero otherwise. Since WSIs at full resolution can be very large ($150,000 \times 150,000$ pixels) and cannot fit into GPU memory, we apply the previously generated masks and then tile each WSI into patches of size $ 1024 \times 1024$ at a spatial resolution of 0.50 micrometers-per-pixel (MPP). Patches capturing less than 40\% of informative tissue area (mean pixel intensity above 200) are discarded, and the rest of the patches are used, both tumour and non-tumour patches. For each patch $p^i_J$ in a WSI $J$, we obtain its $d=2048$-dimensional feature representation $x^i_J \in R^d$ by passing it through a convolutional neural network (CNN) encoder. Any feature extractor can be used to encode patch-level features however we used the RetCCL model~\cite{retccl}.

\subsection{Maximum Mean Discrepancy Kernels}\label{mmdExplanation}
We are interested in defining a (dis)similarity metric between two arbitrary WSIs $X_I$ and $X_J$. To do this we will instead try to solve the analogous problem of defining a distance metric which can then easily be converted to a similarity metric via a transformation (See Equation \ref{eq:transformation}). We first require that this distance metric is valid, it should satisfy the non-negativity, identity, symmetry, and triangle inequality proprieties~\cite{b1}. On top of this, since we are working with WSIs we need the distance metric to be independent of the orientation or shape of tissue, work with arbitrary feature representations, and it should be able to compare WSIs with different numbers of patches. To capture these additional properties we model each WSI as a set of patches since by definition sets are unordered and can have different sizes thus any metrics defined will implicitly be forced to include these assumptions. More specifically, $X_I = \{x^{1}_{I}, x^{2}_{I},...,x^{N_I}_{I} \}$ where $X_I$ is a WSI such that $N_I$ is the number of patches in the image and $x^{m}_{I}$ represents the $m$th patches feature embedding in image $X_I$.

By formulating the problem in terms of sets it is possible to make use of popular distance metrics in the field of set theory. One such example is Maximum Mean Discrepancy (MMD) which was originally proposed as a kernel two-sample test~\cite{kernelTwoSampleTest}. The MMD between two WSIs $X_I$ and $X_J$ is defined as:
\begin{equation}\label{eq:MMD}
MMD(X_I,X_J) = \| \mathbb{E}_{x^i_I \sim X_I}[\varphi(x^i_I)] - \mathbb{E}_{x^j_J \sim X_J}[\varphi(x^j_J)]\|_\mathcal{H}
\end{equation}
where $\mathcal{H}$ is the Reproducing kernel Hilbert space (RKHS) that allows us to perform the change of basis on the patches and $\varphi$ is the feature map $\varphi ; \mathcal{X} \mapsto \mathcal{H}$ that maps into a RKHS~\cite{kernelTwoSampleTest}. Here $\mathcal{X}$ is some space from which $X_I$ and $X_J$ originated. More intuitively, MMD compares different statistical moments between $X_I$ and $X_J$ depending on the specification of $\varphi(\cdot)$. A moment is a quantitative measure that describes specific characteristics of a probability distribution where the first, second, and third moments are mean, standard deviation, and skewness respectively. Formally, the $r^{th}$ moment $\mu_r$ of $X_I$ is defined as $\mu_r = E[X_I^r]$~\cite{momentsDef}. To visualise this consider $\varphi(x) = x$ and $\mathcal{X}=\mathcal{H}={\mathcal{R}^d}$, where ${\mathcal{R}^d}$ is a d-dimensional real space. This will result in:
\begin{equation}\label{eq:MMDExample}
\begin{split}
MMD(X_I,X_J) = & \| \mathbb{E}_{x^i_I \sim X_I}[\varphi(x^i_I)] - \mathbb{E}_{x^j_J \sim X_J}[\varphi(x^j_J)]\|_\mathcal{H} \\
 = & \| \mathbb{E}_{x^i_I \sim X_I}[x^i_I] - \mathbb{E}_{x^j_J \sim X_J}[x^j_J]\|_{\mathcal{R}^d} \\
 = & \| \mu_{X_I} - \mu_{X_J} \|_{\mathcal{R}^d}
\end{split}
\end{equation}
which shows that for this specific transformation, MMD captures the difference in means between the two images. Similarly, we can define $\varphi(x) = x^2$ which allows us to capture the difference between both the mean (first moment) as well as the standard deviation (second moment) of the WSIs.

Based on this principle by setting $\varphi(x) = x^k$ we can compare up to an arbitrary $k^{th}$ moment. However, such an approach is limited because it assumes two images will have a distance of 0 if they share the first $k$ moments. However, this is not correct as the images may only start differing at the $k+1^{th}$ moment. Thus, ideally, we would like to compare an infinite number of moments.

To achieve this we can make use of the kernel trick~\cite{kernelTrick}. This is based on the Moore–Aronszajn theorem and states that a kernel $k(\cdot)$ that is semi-definite positive has an implicit feature representation~\cite{kernelTwoSampleTest}. This is defined as:
\begin{equation}\label{eq:kernelTrick}
k(x^i_I,x^j_J) = \langle x^i_I, x^j_J\rangle = \varphi(x^i_I)^T\varphi(x^j_J).
\end{equation}

Since we have defined $\varphi(\cdot)$ to map to some RKHS it is possible to apply the kernel trick for MMD~\cite{kernelTwoSampleTest}. To do this we first need to rearrange equation \ref{eq:MMD} to be only expressed in terms of dot products and then substitute in the kernel from Equation \ref{eq:kernelTrick}. This results in:
\begin{equation} \label{eq:MMDNew}
\begin{split}
D^{MMD}_{I,J} = MMD^2(X_I,X_J) & = \mathbb{E}_{X_I}[k(x^i_I,x^{i'}_I)] + \mathbb{E}_{X_J}[k(x^j_J,x^{j'}_J)] \\
 & - 2 \mathbb{E}_{X_I,X_j}[k(x^i_I,x^j_J)].
\end{split}
\end{equation}

There is a degree of choice with the exact definition of the kernel (for example Energy or Laplacian~\cite{geomloss}) however, in this case, we use the patch-level Gaussian kernel defined as $k(x^i_I,x^j_J) = e^{\frac{-\| x^i_I-x^j_J \|^2}{4 \sigma^2}}$ where $\sigma$ is the standard deviation (also called blur parameter) of the kernel~\cite{geomloss}. This specifies how large the region of influence of a particular point is, the larger the $\sigma$ the bigger the region. This kernel is semi-definite positive but what's more interesting is that it can be shown that the Gaussian kernel allows us to compare an infinite number of moments and thus MMD will be zero if and only if the distributions are identical~\cite{kernelTwoSampleTest}.

With this, we have defined a pairwise distance metric between any two arbitrary WSIs with the same patch feature embedding. Using this we can generate an $N \times N$ matrix where $N$ is the number of WSIs in the database, $D^{MMD}$. Specifically, $D^{MMD}_{I,J} = MMD^2(X_I,X_J)$ such that $I,J = 1 \ldots N$. This allows us to summarise an entire dataset of WSIs in a single matrix that can fit into computer memory. To get a corresponding symmetric and positive semi-definite Mercer kernel matrix, $K^\gamma \in R^{N\times N}$, from this distance matrix we can apply the transformation: 
\begin{equation}\label{eq:transformation}
    K^{\gamma}=e^{-\gamma D^{MMD}}
\end{equation}
where $K^\gamma_{I,J} \in [0,1]$ is the similarity between $X_I$ and $X_J$ and $\gamma \in \mathbb R_{\ge 0}$ determines the rate at which similarity decreases with increasing distance. Thus for large values of $\gamma$ a small increase in distance will correspond with a big decrease in similarity.

\subsection{Computation of Maximum Mean Discrepancy Kernels}
In line with our previous work, we used the GPU-based GeomLoss library to efficiently compute MMD between WSIs~\cite{geomloss}. In this paper, we use a Gaussian kernel thus to reduce computation the standard deviation for the kernel is set to $\sigma=10$. By parallelising the matrix computation the MMD-dissimilarity matrix for the entire TCGA ($12,186 \times 12,186$ dimensional matrix) took around 2 days to compute with parallelisation. It is important to note this computation was only done once and then the relevant subsections of this kernel were utilised for each task mentioned in Section~\ref{sec2}.

\subsection{Explainability of Maximum Mean Discrepancy Kernels}\label{Explainability}
Since MMD generates a slide-level kernel any models trained with the precomputed kernel are intrinsically unable to generate patch-level predictions since patch-level information has already been condensed during the computation of moments. Thus, it was previously impossible to see which patches were most important for HistoKernel. Explainability is vital in CPath to ensure pathologists can make informed decisions about any models they use as well as to ensure the safety, approval, and acceptance of such models in a clinical setting~\cite{plass2023explainability}. To overcome this limitation, we developed a model-agnostic patch sensitivity metric that tells us how sensitive our predictor is with respect to a certain patch. This method takes inspiration from the field of perturbation-based instance explainability which perturb inputs of a given sample to observe the effects of these perturbations on the model’s output~\cite{ribeiro2016should,robnik2008explaining}. Traditionally, perturbations are applied to the features of a particular instance and any change in the outputs is then attributed to the perturbed feature~\cite{robnik2018perturbation}. Perturbations depend on the type of data but can range from simply increasing or decreasing a numerical feature such as age or adding blur to a region of an image. The changes in the output can then be used to estimate feature importance for a particular sample. In the case of MMD we are not interested in feature importance but rather patch importance. Thus we perturb a given WSI by removing patches to see the impact on the models' output. Specifically, consider some pre-trained predictor, $f: M_{n \times d}(\mathbb{R}) \mapsto \mathbb{R}$, that takes a WSI represented as a set of $n$ d-dimensional patches and generates a real number prediction. Then if we have an arbitrary WSI, $X_I$ with $N_I$ patches, if we passed this into our predictor we would get $f(X_I) = s$ where $s$ is the baseline score generated by the predictor after considering all patches. If we want to see the effect of $j$th patch, $p^j_I$, such that $p^j_I \in 1\dots N_I$ we can remove $p^j_I$ from $X_I$ to see how the score changes. Formally, we can calculate:
\begin{equation}
    \delta_{p_I^j} = f(X^{\neg p_I^j}_I) - f(X_I)
\end{equation}
where $X^{\neg p_I^j}_I$ is WSI $X_I$ without patch $p_I^j$. The magnitude of $\delta_{p_I^j}$ will tell us how impactful the loss of patch $p_I^j$ is to the original prediction which acts as a measure of patch-level importance. We can also use the sign of $\delta_{p_I^j}$ to generate a patch-level prediction. For example, consider the case where we are training a TP53 point mutation classifier where WSIs whose predicted value above some threshold $t$ are classified as mutated and those below as wild type. Now consider we have a test WSI, $X_I$, which is TP53 mutated, equivalently we can say that if $X_I$ is correctly classified it should satisfy $f(X_I) = s > t$. If $\delta_{p_I^j} < 0$ that means the model is less certain that $X_I$ is TP53 mutated after losing $p_I^j$ as  $f(X^{\neg p_I^j}_I)$ was smaller than $f(X_I)$. In other words, patch $p_I^j$ looked like it was TP53 mutated as losing it made the model reduce its confidence that this is a TP53 mutated image. Similarly, if $\delta_{p_I^j} > 0$ we can argue that patch $p_I^j$ resembles TP53 wild type.

These scores can then be normalised, in this paper we use min-max normalisation, for easy visualisation of the WSI patch-level heatmap.

\subsection{Kernel Visualisation}
To perform kernel visualisation we first carried out Uniform Manifold Approximation and Projection for Dimension Reduction (UMAP). This works by constructing a high-dimensional graph representation of the data and then optimising a low-dimensional graph to be as structurally similar as possible~\cite{sainburg2021parametric}. Since the current implementation of UMAP accepts a precomputed square distance matrix we can simply pass in the HistoKernel matrix as input. The two main hyper-parameters of the model are the minimum distance between embedded points, $d_{min}=0.0$, and the size of the local neighbourhood, $s_{local}=100$ (see the documentation for more details~\cite{sainburg2021parametric}). Secondly, we generated a heatmap of the distance matrix. This is done by first converting the distance matrix to a similarity kernel, $K^{MMD}$, with $\gamma$ set to the median of the flattened distance matrix. This standardises all the values between zero and one. Then to get back to a distance matrix, $D^{\prime}$ , we calculate $D^{\prime} = 1 - K^{MMD}$. Finally, we cluster this new matrix using hierarchical clustering using Ward linkage to see which WSIs are clustered together~\cite{sainburg2021parametric}.

\subsection{Drug Sensitivity Prediction}
In order to predict drug sensitivity we made use of support Vector Regression (SVR) with our precomputed kernel. SVRs work similarly to traditional SVMs however the two major differences are that they predict scalar values instead of a binary label and use epsilon-insensitive loss for the loss function. This loss function introduces a new hyper-parameter that is not present in traditional SVMs, $\epsilon$, which sets errors that are within $\epsilon$ distance of the true value to zero. For this experiment, we trained an SVR for each chemical compound separately resulting in 427 models. 

In line with the comparison study, for evaluation we made use of 5-fold cross-validation where we reserve $10\%$ of the training data as a validation set for hyper-parameter optimisation. The metric used for evaluation was Spearman's rank correlation coefficient (SCC) between the ground truth and predicted values. The p-value associated with each SCC was also computed. A Wilcoxon paired signed rank test was also calculated between HistoKernel and the comparative method results~\cite{pairedTTest}. This test is used to establish if two groups of samples are statistically significantly different from one another. A one-sided test was carried out to see if the distribution $d= R_{MMD} - R_{B}$, where $R_{MMD}$ and $R_{B}$ are the results of HistoKernel and comparative method respectively, is stochastically greater than a distribution symmetric about zero. This test was chosen as the Shapiro-Wilk test for normality revealed that both the comparative method and HistoKernel results do not follow a normal distribution ($p \ll 0.01$)~\cite{normalityTest}.

\subsection{Whole Slide Image Retrieval}
\label{methodWSIRet}
In this task for a query WSI, $X_Q$, given that we know the site of origin of $X_Q$ we aim to retrieve the top $k$ most "similar" WSIs. In the comparative method paper, similarity is defined as images from the same cancer sub-type. For example, if $X_Q$ originates from the brain and is of sub-type brain lower grade glioma (LGG) a database of WSIs originating from the brain is searched and the algorithm is considered successful if it returns images of the same sub-type (in this case LGG). To remain fair we will utilise the same definition. 

To perform Whole Slide Image Retrieval we directly query the precomputed kernel. Thus to retrieve the top-k most similar WSIs for an arbitrary query image $X_Q$ using MMD kernels we first generate a similarity kernel of the entire dataset $K^{MMD} =e^{-\gamma D^{MMD}}$. Here $\gamma$ is set to the median of the flattened distance matrix. Then we select the $q^{th}$ row of the matrix, $K^{MMD}_{q*}$. Since $K_{q*}$ captures the similarity between $X_Q$ and every other image in the dataset we just need to sort the row in descending order of similarity to find out which images are most similar to the query. We will also need to exclude the entry from the this sorted row which corresponds to similarity between $X_Q$ with itself since the similarity kernel includes the entire dataset and an image will have a maximum similarity of one with itself, by definition of MMD with a Gaussian kernel~\cite{kernelTwoSampleTest}. 

For performance evaluation, in line with the comparative method, we will be using a majority vote at the top k search results ($mMv@k$) which is given by the formula: 
\begin{equation}
 mMv\text{@}k = \frac{1}{N}\sum_{I=1}^{N}\zeta(y_I,\hat{y}_{I}\lbrack :k \rbrack)
\end{equation}
where $N$ is the number of test WSIs, $y_I$ is the ground truth (the cancer sub-type), $\hat{y}_I[:k]$ are the returned top $k$ most similar images and $\zeta(\cdot)$ is a discriminator function that returns one if the majority label of the predictions is equal to the ground truth and zero otherwise. Here we use $k=5$ so for a given query WSI we retrieve the top five most similar images to the query and if the majority label of these five most similar images matches that of the ground truth of the query the prediction is considered correct. To prevent information leakage we remove all slides from the same patient as $X_Q$ before retrieval. 

\subsection{Point Mutation Prediction}
To predict point mutations we made use of kernelized binary SVMs. Inline with the comparative method we trained a separate model (SVM) for each of the 76 genes. 

For performance evaluation, we used 4-fold cross-validation with $40\%$ of the training data used as a validation set for hyperparameter tuning (both kernel $\gamma$, see Section~\ref{mmdExplanation}, and SVM regularisation parameter)~\cite{pointMutBaseline}. The model's performance was measured by using Area Under the Receiver Operating Characteristic curve (AUC-ROC). In Figure~\ref{pointPredictionBar} part (a) predcitons are binned based on AUC-ROC into weak (AUC-ROC $<$ 0.6), moderate (AUC-ROC 0.6 to 0.7) and strong (AUC-ROC 0.7 to 1.0) predictors. In Fig~\ref{pointPredictionBar} part (b) we filtered out genes where both comparative method and HistoKernel achieved an AUC-ROC less than 0.6, leaving 34 genes. This is because comparing genes for which both predictors are weak is meaningless.

\subsection{Survival Analysis}
\label{sec:methodSurvival}
In order to perform survival analysis we utilised kernelized survival SVMs (KSSVM) with our precomputed kernel. These predict a risk score  $f(X_I)$ for a given patient,$I$, after training a model over a training dataset in the form of $\{(X_I,T_I,\delta_I)|I=1\ldots N_{train}\}$. Each patient is modelled as a tuple comprised of a patient's WSI $X_I$, their disease-specific survival time $T_I$ and a binary event indicator variable $\delta_I \in \{0,1\}$ which shows if the patient has passed away from the recorded cancer or not within a censoring time $T_{censor} = 10$ years. The implementation of KSSVM accepts a pre-computed kernel and has a single hyperparameter $\alpha$ that controls the loss penalty term in its objective function~\cite{fastTrainingSurvivalAnalysis}. We chose, $\alpha=0.0625$ and $\gamma$, for the kernel, is set to the median of the flattened distance matrix. This is done to reduce the number of tunable hyper-parameters as the effect of $\gamma$ is also modulated by the blur parameter of the patch level kernel~\cite{previousMMD}. 

For performance evaluation, in line with the comparative method, we used 5-fold cross-validation stratified with respect to the survival event. In each run we generate the risk scores on the test set which are then used to compute the Concordance-index (C-index)~\cite{cIndex}. This metric ranges from 0.0 to 1.0 and measures the degree of concordance between the relative prediction scores of test patients and their actual survival times. A score of 0.0 indicates an inverted ranking of survival scores whereas 1.0 shows perfect concordance between prediction scores and actual survival time. A score of 0.5 implies no concordance between predicted scores and actual survival times)~\cite{cIndex}. For the statistically significant models we also show the Kaplan-Meier survival curves of the high-risk and low-risk patient groups using the medium of the training set scores as the threshold. The p-value of the log-rank test across the 5 folds is calculated as $2\times median(\{p_r|r = 1\ldots 5\})$ where $p_r$ is the p-value at fold $r$. 

After extracting patch-level predictions using our perturbation-based method (see Methods~\ref{Explainability}) for KIRC WSIs we identified representative patches using K-medoid clustering. Specifically, we form a high risk set of patches by taking the highest scoring patch per patient (randomly picking a single patch in case of draws) then performing k-medoid clustering with $K=25$ to get the 25 most representative high scoring patches. A similar process was applied to find the 25 representative low-scoring patches (by instead taking lowest scoring patches from patients).

\subsection{Multi-Modal Kernels}
\label{sec:methodMultiLearning}
To perform multi modal learning we first have to define our kernels. The first kernel is our precomputed HistoKernel that captures morphological information, $K^{WSI}$. The second is a transcriptomic kernel that is computed from gene expression topics defined by Dawood et al~\cite{dawood2023data}. In this work, each BRCA patient has been characterised by 200 binary variables or topics. The status of each topic variable represents specific gene expression patterns of co-dependent genes. Thus for each patient, $P_I$, we can express their genetic information as a 200 binary feature vector, $\mathbf{t^{P_I}}$: 
\begin{equation}
\mathbf{t^{P_I}} = \begin{bmatrix}
t^{P_I}_1\\
\vdots\\
t^{P_I}_{200}
\end{bmatrix}
\end{equation}
where $t^{P_I}_j\in \{0,1\}$ is the binary status of topic $j$ of patient $I$. To define a topic kernel we can compute the radial basis function between topic representations of two patients as:
\begin{equation}
K^{TOPIC}_{I,J} = e^{-\frac{\lVert t^{P_I} - t^{P_J} \rVert^2}{2 \sigma^2}}
\end{equation} where $\sigma$ is the bandwidth of the RBF kernel function. In this work, for simplicity, we set $\sigma = 10$.

% (-\frac{\lVert t^{P_I} - t^{P_J} \rVert)^2}
We can then utilise the rich theory of Multi Kernel Learning (MKL) in order to combine our set of kernels $K = \{K^{TOPIC}, K^{WSI} \}$. The simplest type of method to combine kernels is Unweighted Multiple Kernel Methods (UMKL) where we perform a specific arithmetic operation to combine kernels without assigning kernel weights~\cite{gonen2011multiple}. The first operation we use is addition where we define a  new kernel $K^{WSI} + K^{TOPIC}$. Addition of kernels is equivalent to concatenating the feature
space representation of the base kernels~\cite{viljanen2022generalized}. Another operation is to multiply the kernels such that $K^{WSI} \times K^{TOPIC}$. This can be thought of as computing the tensor product of of the base kernels feature spaces~\cite{viljanen2022generalized}.

Once these multi-modal kernels are computed they can be used as precomputed kernels for kernelized survival SVMs as shown in Methods~\ref{sec:methodSurvival}. To find the optimum hyperparameters we perform Bayesian optimisation over the validation set. Specifically, we optimise over $\alpha \in [2.0^{-12},0.125]$ which controls the degree of regularisation in the KSSVM function. We then minimise the negative c-index on the validation set added to the chosen $\alpha$. This tries to find hyperparameters that balance a high performance whilst also keeping high regularisation to prevent overfitting.

For performance evaluation, we used 3-fold cross-validation stratified with respect to the survival event while utilizing $20\%$ of the training set for validation. In each run we generate the risk scores on the test set which are then used to compute the C-index and p-value (as shown in Methods~\ref{sec:methodSurvival}).

\section{Conclusion}\label{sec13}
This work explores Maximum Mean Discrepancy-based kernels for various CPath tasks. We have demonstrated the power of HistoKernel for WSI retrieval, cancer drug sensitivity prediction, point mutation classification, survival analysis and multi-modal learning. For these, HistoKernel has outperformed various comparative methods showing its wide versatility and ability to capture meaningful information from WSIs and summarise the data in a very compact manner. Our second major contribution was developing a patch sensitivity metric that provides explainability of slide-level predictions removing a major obstacle that was blocking the use of HistoKernel in a clinical setting. We hope this work paves the way for further exploration of HistoKernel in CPath. Future work will involve testing HistoKernel on datasets other than TCGA, testing for confounding factors and seeing how the model performs in a clinical setting.

\backmatter

% commented
\bmhead{Supplementary information}
Accompanying supplementary is provided as a PDF document.

\section*{Declarations}
\subsection{Funding}
Fayyaz Minhas acknowledges funding from EPSRC EP/W02909X/1. Fayyaz Minhas and Muhammad Dawood report research funding from GlaxoSmithKline outside the submitted work. Piotr Keller acknowledges funding from  EPSRC Doctoral Training Partnership (DTP) at the University of Warwick. Brinder Singh Chohan has no associated funding.
\subsection{Conflict of interest/Competing interests}
Not applicable.
\subsection{Ethics approval and consent to participate}
Not applicable.
\subsection{Consent for publication}
Consent for publication is given by all authors.
\subsection{Data availability}
All WSIs and survival data used in this study are part of the TCGA and are publicly available at \underline{https://www.cancer.gov/ccg/research/genome-sequencing/tcga}. Point mutation data from all genes is publicly available at \underline{https://www.cbioportal.org/31}.
\subsection{Materials availability}
Not applicable.
\subsection{Code availability}
All code is available in GitHub at: \underline{https://github.com/pkeller00/HistoKernel}.
\subsection{Author contribution}
The concept for the study and experimented was developed by P.K. and F.M. The experiments and statistical were implemented by P.K. Assistance with running comparative  models and providing cleaned data was done my M.D. The study was supervised by F.M. B.R analysed patches and WSIs from KIRC, identifying morphological patterns in high and low risk regions.

%%===========================================================================================%%
\bibliography{sn-bibliography}% common bib file

\end{document}